\documentclass[11pt]{article}
\usepackage{graphicx}
\usepackage{subcaption} 
\usepackage{threeparttable}
\usepackage[final]{acl}
\usepackage{times}
\usepackage{latexsym}
\usepackage{fixltx2e}
\usepackage[T1]{fontenc}

\usepackage[utf8]{inputenc}

\usepackage{microtype}
\usepackage{amsmath}
\usepackage{inconsolata}

\usepackage{graphicx}

\usepackage{latexsym}

\usepackage[T1]{fontenc}

\usepackage[utf8]{inputenc}

\usepackage{microtype}

\usepackage{inconsolata}

\usepackage{wrapfig}
\usepackage{graphicx}
\usepackage{amssymb}
\usepackage{amsmath}
\usepackage{subcaption}
\usepackage{subcaption}
\usepackage{booktabs}
\usepackage{multirow}
\usepackage{adjustbox}
\usepackage{colortbl}
\usepackage{subcaption}
\usepackage{placeins}
\usepackage{float}
\usepackage{dblfloatfix}
\usepackage{breqn}
\usepackage{amsthm}
\usepackage{amsmath,amssymb,amsthm}
\theoremstyle{plain}
\newtheorem{theorem}{Theorem}[section]
\newtheorem{lemma}[theorem]{Lemma}         

\theoremstyle{definition}
\newtheorem{definition}[theorem]{Definition}

\theoremstyle{remark}
\newtheorem{remark}[theorem]{Remark}
\title{DoPE: Denoising Rotary Position Embedding}

\author{
  Jing Xiong\textsuperscript{1}\thanks{Equal contribution}, 
  Liyang Fan\textsuperscript{3}\textsuperscript{*}, 
  Hui Shen\textsuperscript{2}, 
  Zunhai Su\textsuperscript{1}, \\
   \textbf{Min Yang}\textsuperscript{3}\thanks{Corresponding author}, 
   \textbf{Lingpeng Kong}\textsuperscript{1}, 
  and  \textbf{Ngai Wong}\textsuperscript{1} \\
  \textsuperscript{1}The University of Hong Kong \textsuperscript{2}University of Michigan, Ann Arbor\\
  \textsuperscript{3}Shenzhen Institute of Advanced Technology, Chinese Academy of Sciences
  \\
 \small{
   \textbf{Contact:} \href{junexiong@connect.hku.hk}{junexiong@connect.hku.hk}
    \vspace{1.0em} \quad
  \textbf{Project:} \url{https://The-physical-picture-of-LLMs.github.io}
 }
}

\begin{document}
\maketitle
\begin{abstract}

Positional encoding is essential for large language models (LLMs) to represent sequence order, yet recent studies show that Rotary Position Embedding (RoPE) can induce massive activation. We investigate the source of these instabilities via a spectral analysis of RoPE, and show that its low-frequency components concentrate structured energy, producing low-rank, over-aligned attention patterns. We theoretically reveal that this low-frequency alignment manifests as activation noise, degrading stability during long-context extrapolation. To mitigate this effect, we introduce Denoising Rotary Position Embedding (\textsc{DoPE}), a training-free method that identifies and suppresses noisy attention heads using \emph{truncated matrix entropy}, then reparameterizes their attention maps with an isotropic Gaussian distribution. Across a range of settings, \textsc{DoPE} improves length extrapolation performance without fine-tuning, increases robustness to perturbations, and boosts both needle-in-a-haystack and many-shot in-context learning tasks. These results suggest that selective positional encoding is key to robust extrapolation.


\end{abstract}
\section{Introduction}

Positional encoding is a core component of large language models (LLMs): it is added to query and key vectors to represent token order and shape interactions among tokens.
Among many approaches~\citep{press2021train,chen2023extending,su2024roformer,peng2023yarn,wang2021kepler}, Rotary Position Embedding (RoPE)~\citep{su2024roformer} is widely used because it encodes relative positions within dot-product attention and often extrapolates well to longer contexts.
While RoPE provides an explicit mechanism for encoding token order, recent work has shown that \emph{causal attention itself}~\citep{gu2024attention,kocher2025nope} implicitly captures positional relationships. Interestingly, this implicit encoding can lead to \emph{massive activations}~\citep{sun2024massive,jin2025massive}, a behavior closely tied to the \emph{attention sink} phenomenon~\citep{xiao2023efficient}.
Yet how explicit positional encodings, especially RoPE, interact with this implicit positional bias and shape massive activations remains poorly understood~\citep{jin2025massive,wu2025emergence}.

Following these observations, recent studies have questioned the necessity of explicit positional encoding, proposing alternatives such as learnable feature maps applied directly to the attention map~\citep{zheng2024dape,zheng2025dape} or even removing positional encoding entirely (NoPE)~\citep{haviv2022transformer,wang2024length,ji2025towards}. These results challenge the necessity of explicit positional encoding and suggest that causal attention may implicitly provide \emph{strong length extrapolation capability} when paired with an appropriate feature map. However, the attention-sink puzzle remains: how the features induce the attention sink, and their underlying mechanism is still unclear. In this work, we investigate how RoPE injects massive activations across heads and introduces structured noise into the attention map, which manifests as the attention sink phenomenon.

We formalize this view by treating the attention map as a noisy feature map through the lens of \textit{truncated matrix entropy}~\citep{xiong2024uncomp}. This perspective lets us detect heads dominated by massive activations and analyze how RoPE contributes to their emergence. We then suppress positional encoding selectively based on \textit{truncated matrix entropy} and reparameterize the corresponding feature maps using an isotropic Gaussian distribution, improving stability in length extrapolation. Specifically, our main contributions are as follows:

\begin{itemize}
\vspace{-2mm}
\item We propose \textsc{DoPE}, a \emph{training-free} denoising scheme for RoPE that selectively suppresses positional encoding and \emph{theoretically} reveal how positional encoding shapes massive activation and attention sink

\vspace{-2mm}
\item We introduce \emph{truncated matrix entropy} to identify heads dominated by massive activations and reparameterize their attention maps with an isotropic Gaussian distribution.

\vspace{-3mm}
\item We show that RoPE’s \emph{low-frequency alignment} induces attention heads with long-range dependency capability, while extrapolative heads are intrinsically low-rank and benefit from preserved positional encoding.
\end{itemize}

\vspace{-3mm}
\section{Related Work}
\vspace{-2mm}
We review length extrapolation methods based on RoPE variants, as well as approaches that extrapolate without explicit positional encodings.

\vspace{-2mm}
\subsection{Length Extrapolation with RoPE}

RoPE~\citep{su2024roformer} is widely adopted because it encodes relative positions directly in dot-product space and often exhibits strong extrapolation. RoPE and its variants are integrated into open-source LLM families, including LLaMA~\citep{touvron2023llama,dubey2024llama}, Qwen~\citep{team2024qwen2,yang2025qwen3}, Mistral~\citep{mistral7b}, and Gemma~\citep{team2024gemma,team2025gemma}.

However, when input sequences exceed the training length~\citep{peng2023yarn,chen2023clex,ding2024longrope}, performance can degrade substantially. This limitation is not unique to RoPE; similar behavior is observed with other relative positional encodings such as ALiBi~\citep{press2021train} and Kerple~\citep{chi2022kerple}.

Notably, several of these extensions modify RoPE at inference time without any training, e.g., by rescaling or interpolating the rotary frequencies. Prior work extends positional encodings in several ways, including interpolation-based~\citep{li2023functional,chen2023positionalinterp} and NTK-based methods~\citep{chen2023clex,peng2023yarn,bloc2023ntkaware,bloc2023ntkawareinterp,emo2023dynamicrope}, which adjust positional scaling or the frequency spectrum to enlarge the effective context. Another line of research~\citep{chen2023clex,ding2024longrope} adopts continuous formulations of positional encodings, modeling them as differential processes to support length extrapolation.

\subsection{Length Extrapolation without Explicit Positional Encoding}

Positional encodings are often viewed as important for sequence awareness and model expressivity~\citep{shaw2018self,yun2019transformers,luo2022your}. Nevertheless, multiple studies~\citep{haviv2022transformer,zuo2024position,kocher2025nope,wu2024role} suggest that causal attention can implicitly capture token order information. The No Positional Encoding (NoPE) approach~\citep{kazemnejad2023impact} argues that the causal mask itself provides sufficient relative position cues, enabling position-aware behavior without explicit positional embeddings. \citet{zuo2024position} further show that such information can emerge through embedding similarity, while \citet{wang2024length} argue that these implicit cues may be insufficient for robust length generalization. This paper therefore examines when positional information should be applied selectively to improve extrapolation.

\begin{figure*}[t]
  \centering
  \includegraphics[width=1.0\linewidth]{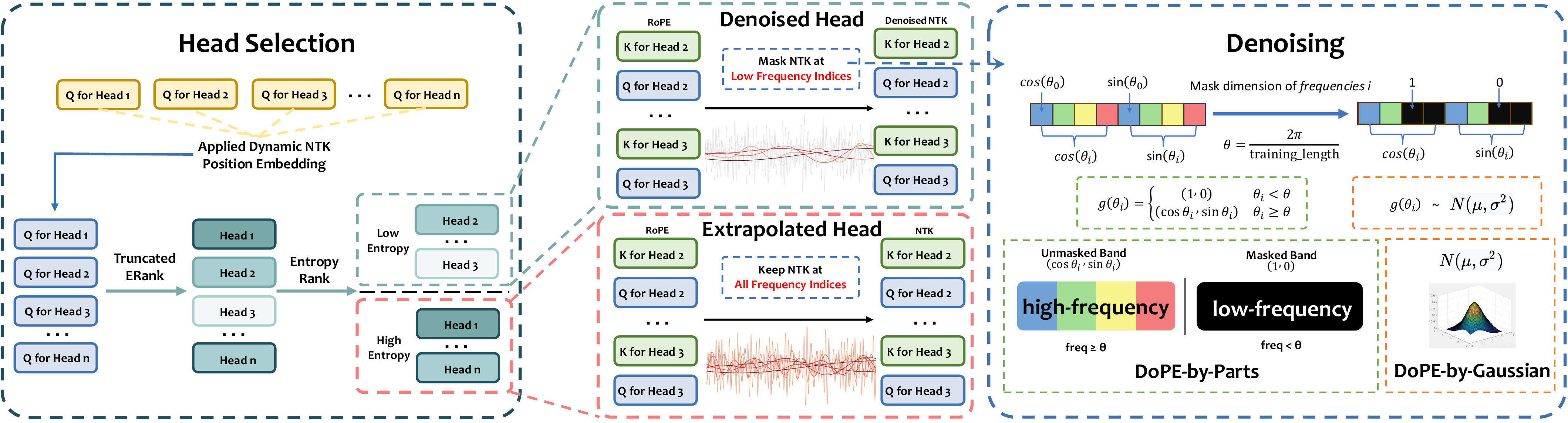}
  \caption{Visualization of DoPE. The blue dashed part illustrates how we select positional encodings for masking.}

  \label{fig:position_id3}
\end{figure*}

\section{Denoising Rotary Position Embedding}

In this section, we first review RoPE and NoPE. 
We then analyze how RoPE induces \textit{attention sink} under the cone constraint, 
which motivates our use of a hybrid architecture combining NoPE and RoPE. Finally, we describe how to identify and remove the corresponding frequency components to mitigate anomalies in attention maps. This process is referred to as \emph{denoising}.

\subsection{Preliminary}
In this section, we briefly review the RoPE and NoPE method separately.

\paragraph{RoPE.}
Let the per-head width be $d_h$. Split a head into $d_h/2$ complex components by pairing dimensions $(2f,2f{+}1)$. 
The frequency band index $f \in \{0, \ldots, d_h/2{-}1\}$ enumerates the $d_h/2$ two-dimensional subspaces, 
each corresponding to a distinct rotation frequency. 
For an integer position $i$ and a frequency schedule $\{\theta_f\}_{f=0}^{d_h/2-1}$ (with base $b{>}1$), 
define the per-band rotation phase $\theta_{i,f}$ and the corresponding $2\times2$ rotation matrix
\begin{equation}
\mathbf{R}(\theta_{i,f}) =
\begin{bmatrix}
\cos \theta_{i,f} & -\sin \theta_{i,f}\\
\sin \theta_{i,f} & \phantom{-}\cos \theta_{i,f}
\end{bmatrix}.
\end{equation}

The full rotation operator is then the block-diagonal matrix
\begin{equation}
\textbf{R}(\theta_i) = \mathrm{diag}\!\big(\mathbf{R}(\theta_{i,0}), \ldots, \mathbf{R}(\theta_{i,d_h/2-1})\big).
\end{equation}
A common choice of frequency schedule is $\theta_f = b^{-2f/d_h}$. For any positions $i,j$, RoPE rotates queries and keys as
\begin{equation}
\textbf{Q}_i^{\mathrm{R}} = \mathbf{R}(\theta_i) \textbf{Q}_i, \qquad
\textbf{K}_j^{\mathrm{R}} = \mathbf{R}(\theta_j) \textbf{K}_j.
\end{equation}
 For $\mathbf{Q}_{i,f}, \mathbf{K}_{j,f} \in \mathbb{R}^{d_h}$ denote the two-dimensional components of 
$\mathbf{Q}_i$ and $\mathbf{K}_j$ obtained 
by pairing dimensions $(2f, 2f{+}1)$,
\begin{equation}
\begin{split}
\big\langle \textbf{Q}_i^{\mathrm{R}} ,\; \textbf{K}_j^{\mathrm{R}} \big\rangle
&= \sum_{f=0}^{d_h/2 - 1} 
\big\langle 
\mathbf{R}(\theta_{i,f}) \mathbf{Q}_{i,f},\;
\mathbf{R}(\theta_{j,f}) \mathbf{K}_{j,f}
\big\rangle\\
&= \sum_{f=0}^{d_h/2 - 1} 
\big\langle 
\mathbf{Q}_{i,f},\;
\mathbf{R}(\theta_{j,f} - \theta_{i,f}) \mathbf{K}_{j,f}
\big\rangle,
\end{split}
\end{equation}
so the attention logits depends on the relative positional offset $(j{-}i)$ 
while preserving the efficiency of the dot product.

\paragraph{NoPE.} In the NoPE method, positional encoding is entirely removed from the attention computation. 
Queries and keys are learned solely from token content without any explicit positional bias. 
Although this avoids the \textit{attention sinks} introduced by RoPE, 
it undoubtedly requires training the model from scratch, which is computationally prohibitive. 
Moreover, when and how RoPE should be transformed into NoPE to prevent attention sinks 
remains theoretically unclear.


\subsection{Spectral Amplification of RoPE Bands}

\label{sec:Spectral Characterization}
\emph{In this section, we present our theoretical contributions.} 
We analyze how the \emph{massive activations} of low-frequency RoPE bands arise through their band-wise Gram matrices, 
providing a theoretical framework for understanding this underlying ``physical picture.''

\paragraph{Cone Constraint.}
Consider projected keys
\begin{equation}
\textbf{K}_f^{\mathrm{R}} = \beta_f\,\mathbf{R}(\theta_f)\,\textbf{K}_f,
\qquad
\beta_f \ge \beta_{\min} > 0,
\end{equation}
where \( \mathbf{R}(\theta_f) \in \mathbb{R}^{2 \times 2} \) is a rotation matrix by phase \( \theta_f \) and \( \beta_{\min} \) denote the minimum scaling factors associated with the query and key frequency bands. The rotation matrix \( \mathbf{R}(\theta_f) \) is used to rotate the band-wise matrix \( \textbf{K}_f \) by an angle \( \theta_f \).

Following the \emph{cone condition} of~\citet{deshpande2014cone}, we define that within a low-frequency band, the RoPE rotations stay within a narrow angular cone. There exists a unit vector \( u \) and a half-angle \( \gamma_K < \tfrac{\pi}{2} \) such that
\begin{equation}
\resizebox{0.85\linewidth}{!}{$
\langle u,\,\mathbf{R}(\theta_f)\textbf{K}_{j,f}^{\mathrm{R}}\rangle
\;\ge\; \|\textbf{K}_{j,f}^{\mathrm{R}}\|\cos\gamma_K
 ,\forall f \in \{1,\dots,d_h/2\}$,
 }
\end{equation}
where operator \( \|\cdot\| \) denotes the Euclidean norm and symbol \( \langle \cdot, \cdot \rangle \) represents the dot product. Intuitively, this means that the phase rotations do not wrap around the circle within the visible context, so all projected queries and keys roughly align in the same direction.

\begin{lemma}[Spectral Amplification]
\label{lem:rope-spectrum}
Under the above cone condition, the band-wise Gram matrix of a sequence with length \( N \)
\begin{equation}
\mathbf{\Sigma}_{j,f}
= \sum_{i=1}^{N}\mathbf{K}_{i,f}^{\mathrm{R}}\,(\mathbf{K}_{j,f}^{\mathrm{R}})^{\!\top}
\end{equation}
captures how the frequency band \( f \) aligns around key position \( j \). Its top eigenvalue is bounded as

\begin{multline}
\lambda_{\max}(\mathbf{\Sigma}_{j,f})
\;\ge\;
N\,\beta_{\min}^{2}\,\|\mathbf{K}_{j,f}^{\mathrm{R}}\|^{2}\cos^{2}\gamma_K,\\
\sigma_{1}(\mathbf{K}_{j,f}^{\mathrm{R}})
\;\ge\;
\beta_{\min}\,\|\mathbf{K}_{j,f}^{\mathrm{R}}\|\,\sqrt{N}\,\cos\gamma_K,
\end{multline}
where \( \lambda_{\max}(\mathbf{\cdot}) \) is the largest eigenvalue, and \( \sigma_1(\cdot) \) is the dominant singular value.
\end{lemma}
\paragraph{Massive Activation.} This lemma characterizes that some stable \emph{directions} are reinforced. As the network depth increases, it leads to the accumulation of large $\ell_2$ norms and resulting in \emph{massive activation}. An analogous result holds for \( \mathbf{Q}_{f}^{\mathrm{R}} \), with parameters \( (\alpha_{\min}, \gamma_Q) \). 

\paragraph{Attention Sink.} We further extend this result to the scenario where the key and query matrices are multiplied. Specifically, for the attention logits submatrix corresponding to frequency band \( f \), we have the following expression:
\begin{equation}
\mathbf{A}_{j,f} = \frac{\mathbf{Q}_{f}^{\mathrm{R}} \, (\mathbf{K}_{j,f}^{\mathrm{R}})^{\top}     }{\sqrt{d}},
\end{equation}
where \( \mathbf{Q}_{f}^{\mathrm{R}} \) and \( \mathbf{K}_{j,f}^{\mathrm{R}} \) are the query matrix and the key representation of the \( j \)-th token, respectively, for frequency band \( f \), and \( d \) is the dimensionality of the query and key vectors. The dominant singular value of the attention matrix for frequency band \( f \) satisfies the following inequality:

\begin{equation}
\resizebox{0.85\linewidth}{!}{$
\sigma_1(\mathbf{A}_{j,f}) \gtrsim 
\frac{\alpha_{\min} \beta_{\min}}{\sqrt{d}} \, 
N \, \|\mathbf{Q}_{f}^{\mathrm{R}}\| \, \|\mathbf{K}_{j,f}^{\mathrm{R}}\| \, 
\cos \gamma_Q \, \cos \gamma_K \, \cos \psi
$}.
\end{equation}
 The parameters \( \gamma_Q \) and \( \gamma_K \) define the half-angles of the cones constraining the directions of the query and key vectors, respectively, 
while \( \psi \) quantifies the angular deviation between the principal directions of \( \mathbf{Q}_{f}^{\mathrm{R}} \) and \( \mathbf{K}_{f}^{\mathrm{R}} \), 
capturing the misalignment of their low-frequency orientations. Complete proofs and matrix inequalities are provided in Appendix~\ref{sec:appendix}. This result formalize how low-frequency RoPE bands \( f \) contribute to attention sinks.

\subsection{Denoising via Truncated Matrix Entropy}

 Recent studies~\citep{jin2025massive,qiao2025q} show that RoPE can induce \emph{outlier channels} in \emph{query} and \emph{key} representations, 
where certain low-frequency bands exhibit large $\ell_2$ norms. However, the $\ell_2$ norm captures only \textit{magnitude}, missing the \textit{directional} anomalies. In this section, we demonstrate how truncated matrix entropy~\citep{xiong2024uncomp} can be used to capture the \emph{Spectral Amplification} effect described in Lemma~\ref{lem:rope-spectrum}.

\paragraph{Truncated Matrix Entropy.}

Following~\citet{xiong2024uncomp}, we define the truncated matrix entropy for attention head \( h \) as:
\begin{equation}
\mathcal{H}_{h}^{r} = \frac{1}{r} \sum_{i=1}^{r} \lambda_i \log \lambda_i,
\end{equation}
where \( \lambda_i \) are the \( i \)-th largest singular values of the Gram matrix \( \mathbf{\Sigma}_{h} \), 
and \( r \) denotes the number of singular values considered. This formulate captures the contribution of the top \( r \) singular values to the entropy of the parameter matrix, such as the key or query matrix, which allows us to assess the \emph{effective rank} of the attention head.

\paragraph{Head Selection.}

We define two types of heads based on their entropy. 
Heads with low matrix truncated entropy are identified as \emph{denoised heads}, 
while the others are treated as \emph{extrapolative heads} that follow standard dynamic-NTK extrapolation~\citep{emo2023dynamicrope}. 
We selected the following heads as the \emph{denoised heads}:
\begin{equation}
m_h = \mathbf{1}\!\big[\mathcal{H}_{h}^{\,r}\ge \tau\big],
\end{equation}
where $\tau$ is a quantile threshold. 
Only heads with $m_h\!=\!0$ (low-entropy spectra) undergo denoising.

\paragraph{\textsc{DoPE}-by-parts.}

Recall that under the cone condition, low-frequency RoPE bands correspond to small phase increments $\psi$, yielding a narrow angular spread. 
In practice, we approximate this low-frequency region by a phase threshold $\theta = 2\pi/L$, such that bands with $\theta_f \le \psi$ are considered to lie within the cone. For selected \emph{denoised heads}, denoising acts per frequency band:

\begin{equation}
m_{h,f} = \mathbf{1}\!\big[\theta_f \le \psi\big],
\qquad
\psi = \frac{2\pi}{L},
\end{equation}

where $L$ is the training length, and yielding

\begin{equation}
\mathbf{Q}^{\mathrm{R,D}}_{h}
= \sum_{f=1}^{d_h/2}
m_{h,f}
\mathbf{Q}^{\mathrm{R}}_{h,f},
\end{equation}

\begin{equation}
\mathbf{K}^{\mathrm{R,D}}_{h}
= \sum_{f=1}^{d_h/2}
m_{h,f}
\mathbf{K}^{\mathrm{R}}_{h,f}.
\end{equation}
This operation removes the corresponding low-frequency components $f$ from the query and key matrices of head $h$.

\paragraph{\textsc{DoPE}-by-all.}

We apply head-level positional encoding masking to the selected denoised heads:

\begin{equation}
\mathbf{K}_{h}^{\mathrm{R,D}} = m_h\,\mathbf{K}_{h}^{\mathrm{R}}, \qquad
\mathbf{Q}_{h}^{\mathrm{R,D}} = m_h\,\mathbf{Q}_{h}^{\mathrm{R}}.
\end{equation}




\paragraph{\textsc{DoPE}-by-Gaussian.}

Alternatively, the positional encodings of denoised heads are fully masked and then replaced with:
\begin{equation}
\begin{aligned}
\mathbf{K}_{h}^{\mathrm{R,D}} &=
(1 - m_h)\,\boldsymbol{\epsilon}_{K,h}\mathbf{K}_{h}^{\mathrm{R}}, \\
\mathbf{Q}_{h}^{\mathrm{R,D}} &=
(1 - m_h)\,\boldsymbol{\epsilon}_{Q,h}\mathbf{Q}_{h}^{\mathrm{R}},
\end{aligned}
\end{equation}

where $\boldsymbol{\epsilon}_{K,h}, \boldsymbol{\epsilon}_{Q,h} \sim \mathcal{N}(0,\sigma^2\mathbf{I})$. This can be viewed as reparameterizing the attention map using a isotropic Gaussian distribution.

\begin{table*}[!t]
\centering
\caption{Summary of denoising configurations and results. \textit{Indicator} specifies whether matrix entropy is computed from \emph{Query} or \emph{Key} representations for head selection. \emph{Entropy Type} is vanilla matrix entropy or Trunc-$r$ ($\mathcal{H}_{h}^{r}$ using the top-$r$ singular values). \# Heads is the number of selected heads. \emph{Criterion} indicates when entropy is computed: pre\_ntk (before NTK scaling), ntk (after NTK positional encoding), or post\_rope (after RoPE). Sort Order defines masking: \textbf{ASC} removes lower-entropy heads, while \textbf{DESC} removes higher-entropy heads. Results are reported at 24{,}756 (24k) and 65{,}536 (64k) tokens.}

\label{tab:detailed_results_filtered}
\begin{adjustbox}{max width=\textwidth}
\begin{tabular}{l c c c c c c c c c}
\toprule
\textbf{Method} & \textbf{Indicator} & \textbf{Entropy Type} & \textbf{\# Heads} & \textbf{Criterion} & \textbf{Sort Order} & \textbf{Noisy (24k)} & \textbf{Original (24k)} & \textbf{Noisy (64k)} & \textbf{Original (64k)} \\
\midrule
Dynamic NTK & -- & -- & -- & -- & -- & 75.417 & 91.896 & 40.417 & 60.938 \\
Dual Chunk Attention & -- & -- & -- & -- & -- & 77.053 &	87.896	 &55.792	 &66.438 \\
Positional Interpolatio & -- & -- & -- & -- & -- & 14.583 &	26.417	 &9.479	 &11.771 \\
\midrule
\textsc{DoPE}-by-Gaussian  & Query & Vanilla & 5 & post\_ntk\_query & DESC & 62.521 & \textbf{94.938} & 23.208 & 36.813 \\
\textsc{DoPE}-by-Gaussian & Key & Trunc-32 & 3 & post\_ntk\_key & ASC & \textbf{84.354} & 94.396 & 40.875 & 60.896 \\
\textsc{DoPE}-by-Gaussian & Key & Trunc-16 & 5 & pre\_ntk\_key & ASC & 77.417 & 93.708 & 40.604 & 60.313 \\
\textsc{DoPE}-by-Gaussian & Query & Trunc-16 & 5 & pre\_ntk\_query & ASC & 77.104 & 93.563 & 25.521 & 46.813 \\
\textsc{DoPE}-by-Gaussian & Key & Trunc-16 & 3 & pre\_ntk\_key & ASC & 77.438 & 93.125 & 41.271 & 60.021 \\
\textsc{DoPE}-by-Gaussian & Key & Trunc-8 & 1 & post\_ntk\_key & DESC & 75.250 & 92.229 & \textbf{45.667} & 64.042 \\
\textsc{DoPE}-by-Gaussian & Key & Trunc-4 & 3 & post\_ntk\_key & DESC & 65.833 & 89.354 & 45.375 & 61.979 \\
\textsc{DoPE}-by-Gaussian & Key & Vanilla & 2 & post\_ntk\_key & DESC & 73.229 & 90.188 & 44.229 & 64.292 \\
\textsc{DoPE}-by-Gaussian & Query & Trunc-1 & 5 & post\_ntk\_query & ASC & 75.167 & 92.938 & 42.208 & \textbf{70.083} \\
\textsc{DoPE}-by-Gaussian & Query & Trunc-1 & 3 & post\_ntk\_query & ASC & 72.583 & 89.688 & 41.479 & 69.438 \\
\textsc{DoPE}-by-Gaussian & Query & Vanilla & 5 & post\_ntk\_query & ASC & 44.833 & 76.188 & 44.042 & 65.854 \\
\midrule
\textsc{DoPE}-by-parts  & Key & Trunc-32 & 30 & post\_rope\_key & ASC & 76.229 & \textbf{93.063} & 40.312 & 60.375 \\
\textsc{DoPE}-by-parts & Query & Trunc-32 & 25 & post\_ntk\_query & ASC & \textbf{76.604} & 93.042 & 40.458 & 61.917 \\
\textsc{DoPE}-by-parts & Key & Trunc-32 & 30 & post\_ntk\_key & ASC & 76.458 & 92.875 & 40.771 & 61.333 \\
\textsc{DoPE}-by-parts & Key & Trunc-32 & 20 & post\_ntk\_key & ASC & 76.042 & 92.854 & 40.188 & 60.625 \\
\textsc{DoPE}-by-parts & Key & Trunc-32 & 25 & post\_ntk\_key & ASC & 76.104 & 92.771 & 40.021 & 61.083 \\

\textsc{DoPE}-by-parts & Query & Trunc-16 & 2 & post\_ntk\_query & DESC & 75.438 & 92.354 & \textbf{42.729} & 60.729 \\
\textsc{DoPE}-by-parts & Query & Trunc-8 & 2 & post\_ntk\_query & DESC & 75.229 & 91.771 & 42.521 & 61.104 \\
\textsc{DoPE}-by-parts & Query & Trunc-8 & 3 & post\_ntk\_query & DESC & 75.271 & 92.146 & 42.438 & 59.583 \\
\textsc{DoPE}-by-parts & Query & Trunc-32 & 3 & post\_rope\_query & ASC & 74.500 & 92.125 & 40.313 & \textbf{62.208} \\
\textsc{DoPE}-by-parts & Query & Vanilla & 3 & post\_rope\_query & ASC & 74.125 & 92.479 & 40.125 & 62.146 \\
\textsc{DoPE}-by-parts & Query & Trunc-32 & 5 & post\_ntk\_query & DESC & 75.438 & 91.958 & 40.938 & 62.125 \\
\midrule
\textsc{DoPE}-by-all  & Key & Trunc-32 & 3 & post\_ntk\_key & ASC & \textbf{81.958} & \textbf{93.833} & 40.917 & 61.271 \\
\textsc{DoPE}-by-all & Key & Trunc-16 & 3 & post\_rope\_key & DESC & 65.958 & 93.771 & 35.354 & 61.063 \\
\textsc{DoPE}-by-all & Key & Trunc-16 & 3 & pre\_ntk\_key & ASC & 76.583 & 93.729 & 41.354 & 57.833 \\
\textsc{DoPE}-by-all & Key & Vanilla & 3 & post\_ntk\_key & DESC & 75.625 & 93.271 & 39.729 & 58.021 \\
\textsc{DoPE}-by-all & Query & Vanilla & 3 & pre\_ntk\_query & ASC & 73.542 & 93.250 & 39.333 & 63.146 \\

\textsc{DoPE}-by-all & Key & Trunc-8 & 1 & post\_ntk\_key & DESC & 74.917 & 92.000 & \textbf{46.000} & 63.625 \\
\textsc{DoPE}-by-all & Key & Trunc-4 & 3 & post\_ntk\_key & DESC & 65.958 & 89.813 & 45.292 & 62.646 \\
\textsc{DoPE}-by-all & Query & Trunc-1 & 2 & post\_ntk\_query & DESC & 75.104 & 92.354 & 44.292 & 64.146 \\
\textsc{DoPE}-by-all & Query & Trunc-1 & 5 & post\_ntk\_query & ASC & 75.000 & 92.917 & 42.729 & \textbf{70.083} \\
\textsc{DoPE}-by-all & Query & Trunc-1 & 3 & post\_ntk\_query & ASC & 73.104 & 90.063 & 41.646 & 69.708 \\
\textsc{DoPE}-by-all & Query & Trunc-1 & 3 & post\_rope\_query & DESC & 46.771 & 87.521 & 27.000 & 69.104 \\
\bottomrule
\end{tabular}
\end{adjustbox}
\end{table*}

\begin{table*}[!t]
\centering
\caption{Summary of denoising configurations and results on Qwen2.5-Math-7B for Many-Shot In-Context Learning extrapolation (4K$\rightarrow$16K). We evaluate (1) Needle Insertion, where the problem is inserted into the ICL haystack at one of four depths (beginning, 1/3, 2/3, end), and (2) Skip Needle, a no-insertion baseline. \textit{Indicator} specifies whether denoising is applied to Query or Key representations. \emph{Entropy Type} is Vanilla (matrix entropy $\mathcal{H}_{h}$) or Trunc-$r$ ($\mathcal{H}_{h}^{r}$ with threshold $r$). \# Heads is the number of selected heads. \emph{Criterion} indicates when entropy is computed: pre\_ntk, ntk, or post\_rope. Sort Order specifies the selection direction: \textbf{DESC} (highest entropy) or \textbf{ASC} (lowest entropy). Results report accuracy on 100 sampled MATH problems (400 total configurations across insertion positions).}
\label{tab:qwen_math_extrapolation}
\vspace{-2mm}
\small  
\begin{adjustbox}{max width=\textwidth}
\begin{tabular}{l c c c c c c c c c}
\toprule
\textbf{Method} & \textbf{Indicator} & \textbf{Entropy Type} & \textbf{\# Heads} & \textbf{Criterion} & \textbf{Sort Order} & \textbf{Needle Insert (8K)} & \textbf{Skip Needle (8K)} & \textbf{Needle Insert (16K)} & \textbf{Skip Needle (16K)} \\
\midrule
Zero-shot Baseline & -- & -- & -- & -- & -- & 0.430 & 0.430 & 0.430 & 0.430 \\
Many-shot Baseline & -- & -- & -- & -- & -- & 0.373 & 0.370 & 0.240 & 0.230 \\
Dual Chunk Attention	& -- & -- & -- & -- & -- &0	&0.01	&0	&0 \\
Positional Interpolation	& -- & -- & -- & -- & -- &0	&0.01	&0	&0\\
\midrule
\textsc{DoPE}-by-Gaussian & Query & Trunc-1 & 1 & post\_ntk\_query & ASC & \textbf{0.393} & 0.410 & 0.228 & 0.250 \\
\textsc{DoPE}-by-Gaussian & Query & Trunc-16 & 1 & post\_ntk\_query & ASC & 0.380 & 0.360 & 0.225 & 0.250 \\
\textsc{DoPE}-by-Gaussian & Query & Trunc-1 & 3 & post\_ntk\_query & ASC & 0.375 & 0.370 & 0.238 & 0.220 \\
\textsc{DoPE}-by-Gaussian & Query & Trunc-4 & 5 & post\_ntk\_query & ASC & 0.375 & \textbf{0.440} & 0.225 & 0.190 \\
\textsc{DoPE}-by-Gaussian & Query & Trunc-1 & 5 & post\_ntk\_query & ASC & 0.318 & 0.440 & 0.238 & 0.220 \\
\textsc{DoPE}-by-Gaussian & Query & Trunc-4 & 3 & post\_ntk\_query & ASC & 0.358 & 0.430 & 0.223 & 0.210 \\
\textsc{DoPE}-by-Gaussian & Query & Trunc-1 & 2 & post\_ntk\_query & ASC & 0.345 & 0.380 & \textbf{0.258} & 0.240 \\
\textsc{DoPE}-by-Gaussian & Query & Full & 1 & post\_ntk\_query & DESC & 0.388 & 0.400 & 0.258 & 0.230 \\
\textsc{DoPE}-by-Gaussian & Query & Full & 3 & post\_ntk\_query & DESC & 0.370 & 0.340 & 0.255 & \textbf{0.270} \\
\textsc{DoPE}-by-Gaussian & Query & Trunc-16 & 3 & post\_ntk\_query & ASC & 0.355 & 0.420 & 0.248 & 0.260 \\
\midrule
\textsc{DoPE}-by-parts  & Query & Trunc-1 & 1 & post\_ntk\_query & ASC & \textbf{0.388} & 0.410 & 0.230 & 0.250 \\
\textsc{DoPE}-by-parts  & Query & Trunc-16 & 2 & post\_ntk\_query & ASC & 0.380 & 0.330 & 0.245 & \textbf{0.260} \\
\textsc{DoPE}-by-parts  & Query & Trunc-4 & 5 & post\_ntk\_query & ASC & 0.368 & 0.390 & 0.220 & 0.260 \\
\textsc{DoPE}-by-parts  & Query & Trunc-4 & 3 & post\_ntk\_query & ASC & 0.360 & \textbf{0.420} & 0.240 & 0.230 \\
\textsc{DoPE}-by-parts  & Query & Trunc-8 & 3 & post\_ntk\_query & ASC & 0.363 & 0.390 & 0.220 & 0.180 \\
\textsc{DoPE}-by-parts  & Query & Trunc-1 & 5 & post\_ntk\_query & ASC & 0.355 & 0.350 & 0.245 & 0.240 \\
\textsc{DoPE}-by-parts  & Query & Trunc-16 & 5 & post\_ntk\_query & ASC & 0.365 & 0.380 & 0.243 & 0.260 \\
\textsc{DoPE}-by-parts  & Query & Full & 1 & post\_ntk\_query & DESC & 0.375 & 0.350 & 0.245 & 0.240 \\
\textsc{DoPE}-by-parts  & Query & Full & 2 & post\_ntk\_query & DESC & 0.400 & 0.380 & \textbf{0.258} & 0.230 \\
\textsc{DoPE}-by-parts  & Query & Full & 3 & post\_ntk\_query & DESC & 0.388 & 0.390 & 0.258 & 0.250 \\
\midrule
\textsc{DoPE}-by-all & Query & Trunc-1 & 1 & post\_ntk\_query & ASC & \textbf{0.395} & 0.430 & 0.235 & 0.240 \\
\textsc{DoPE}-by-all & Query & Trunc-4 & 2 & post\_ntk\_query & ASC & 0.383 & 0.390 & 0.215 & 0.240 \\
\textsc{DoPE}-by-all & Query & Trunc-8 & 2 & post\_ntk\_query & ASC & 0.383 & 0.390 & 0.225 & 0.220 \\
\textsc{DoPE}-by-all & Query & Trunc-1 & 5 & post\_ntk\_query & ASC & 0.338 & \textbf{0.480} & 0.243 & 0.220 \\
\textsc{DoPE}-by-all & Query & Trunc-1 & 3 & post\_ntk\_query & ASC & 0.353 & 0.440 & 0.258 & 0.210 \\
\textsc{DoPE}-by-all & Query & Trunc-4 & 5 & post\_ntk\_query & ASC & 0.375 & 0.440 & 0.220 & 0.200 \\
\textsc{DoPE}-by-all & Query & Trunc-8 & 3 & post\_ntk\_query & ASC & 0.375 & 0.440 & 0.205 & 0.190 \\
\textsc{DoPE}-by-all & Query & Trunc-16 & 5 & post\_ntk\_query & ASC & 0.360 & 0.360 & \textbf{0.263} & 0.240 \\
\textsc{DoPE}-by-all & Query & Full & 3 & post\_ntk\_query & DESC & 0.393 & 0.350 & 0.258 & 0.210 \\
\textsc{DoPE}-by-all & Query & Trunc-1 & 2 & post\_ntk\_query & ASC & 0.363 & 0.380 & 0.243 & 0.250 \\
\textsc{DoPE}-by-all & Query & Trunc-16 & 1 & post\_ntk\_query & ASC & 0.365 & 0.370 & 0.228 & \textbf{0.250} \\
\textsc{DoPE}-by-all & Query & Trunc-16 & 3 & post\_ntk\_query & ASC & 0.353 & 0.340 & 0.253 & 0.240 \\
\bottomrule
\end{tabular}
\end{adjustbox}
\end{table*}

\begin{table*}[!t]
\centering
\vspace{-1mm}
\caption{Ablation study: Performance on 64k extrapolation using attention heads selected at different sequence lengths. Each configuration uses heads identified from sequences of length 24k, 32k, 48k, 56k, and 64k, then evaluates on the 64k task under both Noisy and Original conditions. }
\vspace{-2mm}
\label{tab:ablation_head_selection}
\begin{adjustbox}{max width=\textwidth}
\begin{tabular}{l c c c c c c c c c c c c c c c}
\toprule
& & & & & & \multicolumn{2}{c}{\textbf{24k Heads}} & \multicolumn{2}{c}{\textbf{32k Heads}} & \multicolumn{2}{c}{\textbf{48k Heads}} & \multicolumn{2}{c}{\textbf{56k Heads}} & \multicolumn{2}{c}{\textbf{64k Heads}} \\
\cmidrule(lr){7-8} \cmidrule(lr){9-10} \cmidrule(lr){11-12} \cmidrule(lr){13-14} \cmidrule(lr){15-16}
\textbf{Method} & \textbf{Indicator} & \textbf{Entropy Type} & \textbf{\# Heads} & \textbf{Criterion} & \textbf{Sort Order} & \textbf{Noisy} & \textbf{Original} & \textbf{Noisy} & \textbf{Original} & \textbf{Noisy} & \textbf{Original} & \textbf{Noisy} & \textbf{Original} & \textbf{Noisy} & \textbf{Original} \\
\midrule
Dynamic NTK & -- & -- & -- & -- & -- & 40.417 & 60.938 & 40.417 & 60.938 & 40.417 & 60.938 & 40.417 & 60.938 & 40.417 & 60.938 \\
\midrule
\textsc{DoPE}-by-Gaussian & Key & Trunc-8 & 1 & post\_ntk\_key & DESC & 40.896 & 62.438 & 40.417 & 63.125 & 28.666 & 60.104 & 28.666 & 60.104 & \textbf{45.667} & \textbf{64.042} \\
\textsc{DoPE}-by-Gaussian & Query & Trunc-1 & 5 & post\_ntk\_query & ASC & 35.667 & 56.708 & 30.354 & 61.979 & \textbf{43.604} & 69.166 & 38.020 & 69.854 & 42.208 & \textbf{70.083} \\
\textsc{DoPE}-by-parts & Query & Trunc-16 & 2 & post\_ntk\_query & DESC & 41.792 & 61.271 & 41.479 & 65.020 & 41.479 & 65.020 & 41.479 & \textbf{65.020} & \textbf{42.729} & 60.729 \\
\textsc{DoPE}-by-parts & Query & Trunc-32 & 3 & post\_rope\_query & ASC & 40.313 & 61.688 & 39.333 & 66.979 & 39.333 & 66.979 & 39.333 & \textbf{66.979} & \textbf{40.313} & 62.208 \\
\textsc{DoPE}-by-all & Key & Trunc-8 & 1 & post\_ntk\_key & DESC & 40.625 & 62.208 & 40.541 & \textbf{65.229} & 29.604 & 59.979 & 29.604 & 59.979 & \textbf{46.000} & 63.625 \\
\textsc{DoPE}-by-all & Query & Trunc-1 & 5 & post\_ntk\_query & ASC & 37.063 & 61.292 & 32.687 & 65.000 & \textbf{43.000} & \textbf{75.187} & 40.458 & 73.812 & 42.729 & 70.083 \\

\bottomrule
\end{tabular}
\vspace{-12mm}
\end{adjustbox}
\end{table*}

\vspace{-2mm}
\section{Experiment}
\vspace{-2mm}
\subsection{Experimental Setup}
\vspace{-1mm}
The ``needle-in-a-haystack'' (NIH) synthesis task benchmarks long-context retrieval by placing a sparse ``needle'' at different depths and measuring recall. It also allows controlled noise injection (e.g., special tokens). We evaluate two conditions: \textit{original setups} and \textit{noisy setups}.

\vspace{-2mm}
\paragraph{Original Setups.} We insert the needle at various positions under context lengths of 24K and 64K tokens to measure retrieval performance and the lost-in-the-middle effect.

\vspace{-3mm}
\paragraph{Noisy Setups.} Under the same context lengths, we insert attention-sink tokens (e.g., a start-of-sequence symbol) near the needle to test robustness under controlled perturbations and relate performance to \emph{attention sinks} and matrix entropy.

\vspace{-2mm}
\paragraph{Many-shot In-context Learning.} For many-shot in-context learning (MICL)~\citep{agarwal2024many}, we evaluate standard MICL and its NIH variant at 8K and 16K context lengths. Data are sampled from the MATH dataset~\citep{hendrycksmath2021}.

\begin{figure*}[t]
  \centering
  \begin{subfigure}{0.495\textwidth}
    \centering
    \includegraphics[width=\linewidth]{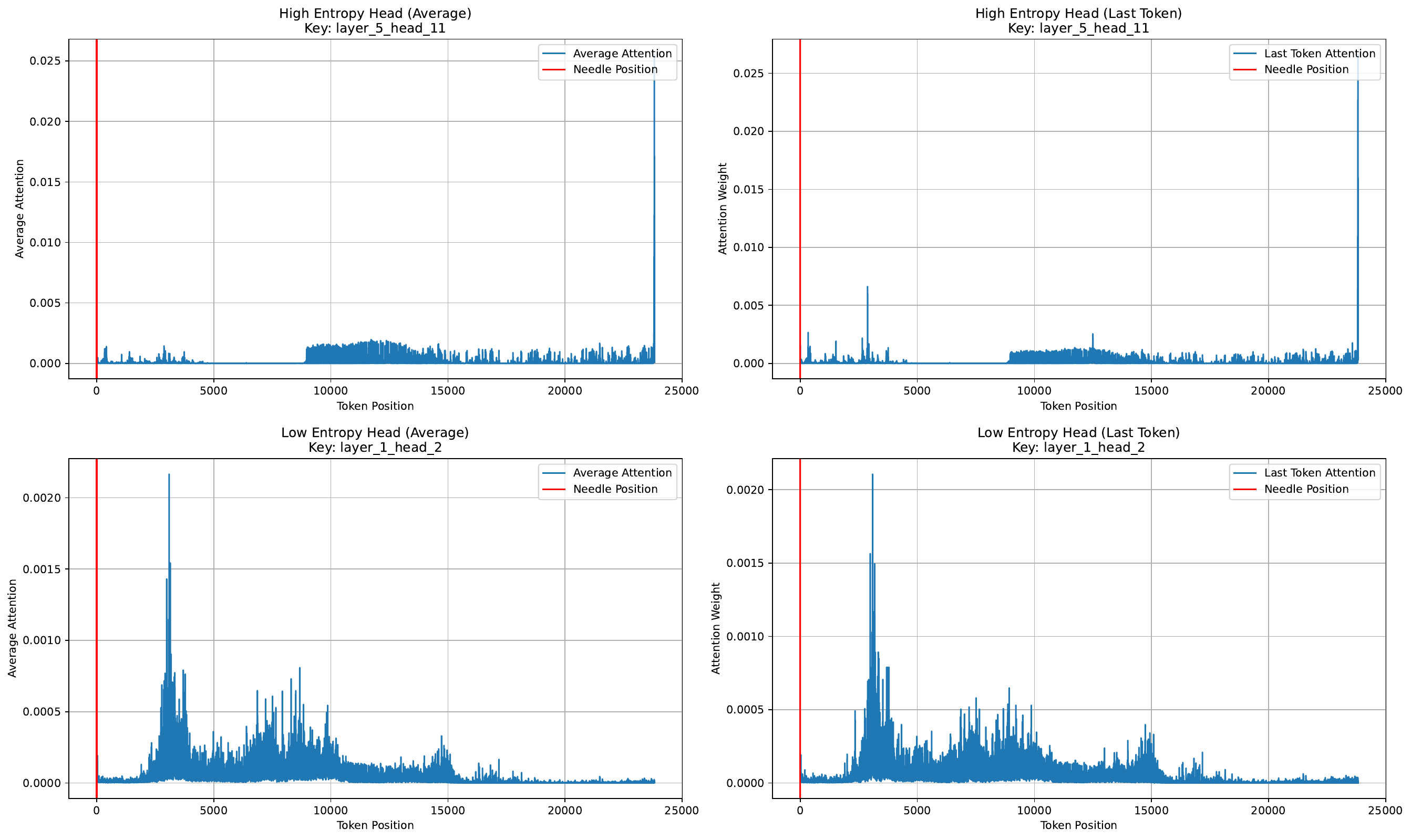}
    \caption{DoPE by Vanilla Matrix Entropy.}
    \label{fig:attn_all_head}
  \end{subfigure}
  \hspace{-0.3em} 
  \begin{subfigure}{0.495\textwidth}
    \centering
    \includegraphics[width=\linewidth]{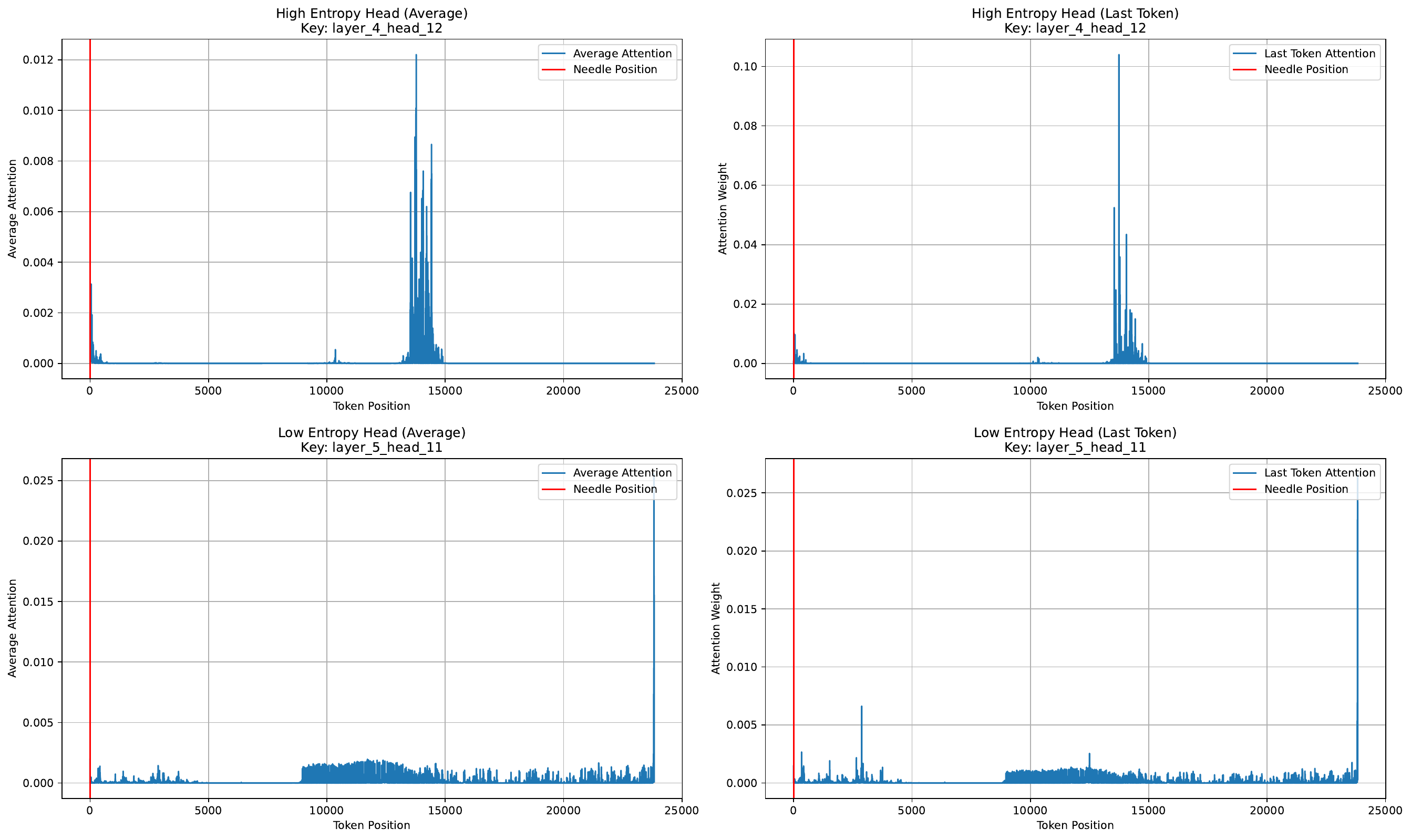}
    \caption{DoPE by Truncated Matrix Entropy.}
    \label{fig:attn_top16_head}
  \end{subfigure}
  \vspace{-3mm}
  \caption{Comparison of attention distribution across all heads and top-16 heads.}
  \vspace{-5mm}
  \label{fig:attn_compare}
\end{figure*}
\vspace{-2mm}

\paragraph{Hyperparameters of Head Selection.}
We select 1--32 heads based on either ascending or descending scores. We use calibration data matched in length to the test data to precompute \textit{matrix entropy}. Entropy can be computed at three stages of the forward pass, each isolating a different positional-effect factor: (1) \textit{pre-NTK}, on projected query/key representations before positional encoding (no PE); (2) \textit{post-NTK}, after Dynamic-NTK scaling of the RoPE base frequency (frequency-scaling effect); and (3) \textit{post-RoPE}, after applying the RoPE rotation (full PE effect).

Entropy can be computed on \textit{query} or \textit{key} representations; in practice, the \textit{query} matrix~\citep{tang2024quest} better captures head characteristics. This yields six criteria (3 stages $\times$ 2 components), plus an option to compute entropy jointly on query and key. We denote configurations as \textit{Criterion}; for example, \texttt{post\_ntk\_query} computes entropy on query representations after NTK scaling.

\subsection{Main results}

We conducted experiments under two settings: \textit{original setups} and \textit{noisy setups}, with the results summarized in Table~\ref{tab:detailed_results_filtered}. Our findings are summarized as follows: \textit{(i)} The model exhibits a sharp performance degradation after introducing attention sink tokens. \textit{(ii)} Under the shorter context setting (24k tokens), \textsc{DoPE}-by-Gaussian achieves its best performance, improving from the \textbf{75.417} baseline to \textbf{84.354}. The inclusion of a \textit{Gaussian} distribution generally promotes isotropy in representations, which usually increases the discriminability of token representations in the denoised head, allowing the model to focus on a few important tokens. \textit{(iii)} \textit{Truncated matrix entropy} and \textit{(vallina) matrix entropy} exhibit distinctly different patterns. For the truncated variant, we sort values in descending order and prune the low-entropy heads; for the matrix entropy, we sort in ascending order and prune the high-entropy heads. Both strategies perform well, but truncated matrix entropy typically achieves better results. \textit{(iv)} In extremely sparse regimes—for example, with a 64K context length—using the truncated matrix entropy with $r=1$ (which can be regarded as equivalent to the \textit{spectral norm}, i.e., $\sigma_{max}(\mathbf{\Sigma_h}    )$) yields the best results. This indicates that the sparser the setting, the sharper the singular value distribution becomes. \textit{(v)} As shown in Table~\ref{tab:ablation_head_selection}, cross-length calibration generally performs worse than same-length calibration.

\begin{figure*}[!t]
  \centering
  \begin{minipage}{0.48\textwidth}
    \centering
    \includegraphics[width=\linewidth]{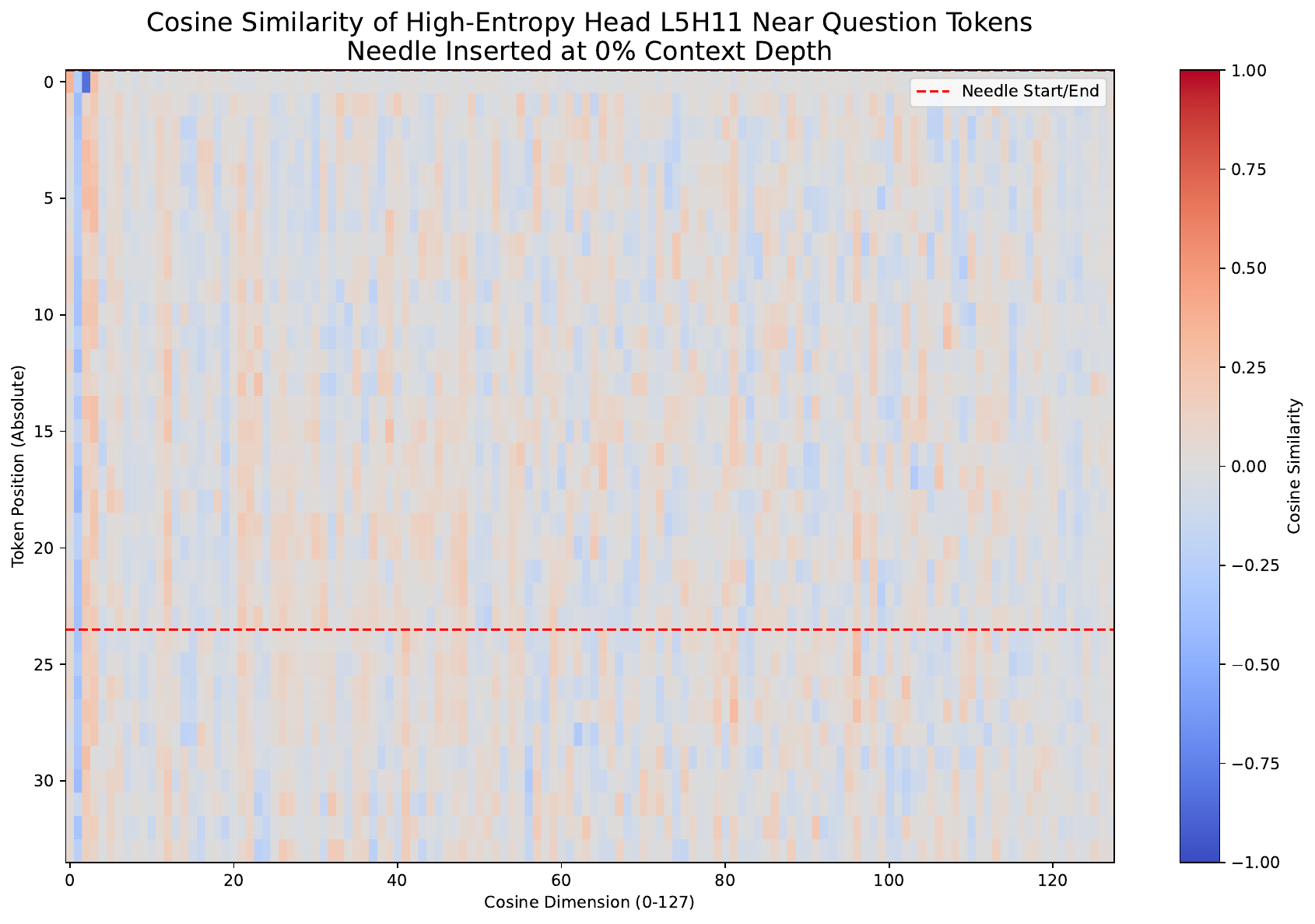}
    \caption{High matrix entropy head (Layer 5, Head 11)}
    \vspace{-3mm}
    \label{fig:high_entropy_1}
  \end{minipage}
  \hfill
  \begin{minipage}{0.48\textwidth}
    \centering
    \includegraphics[width=\linewidth]{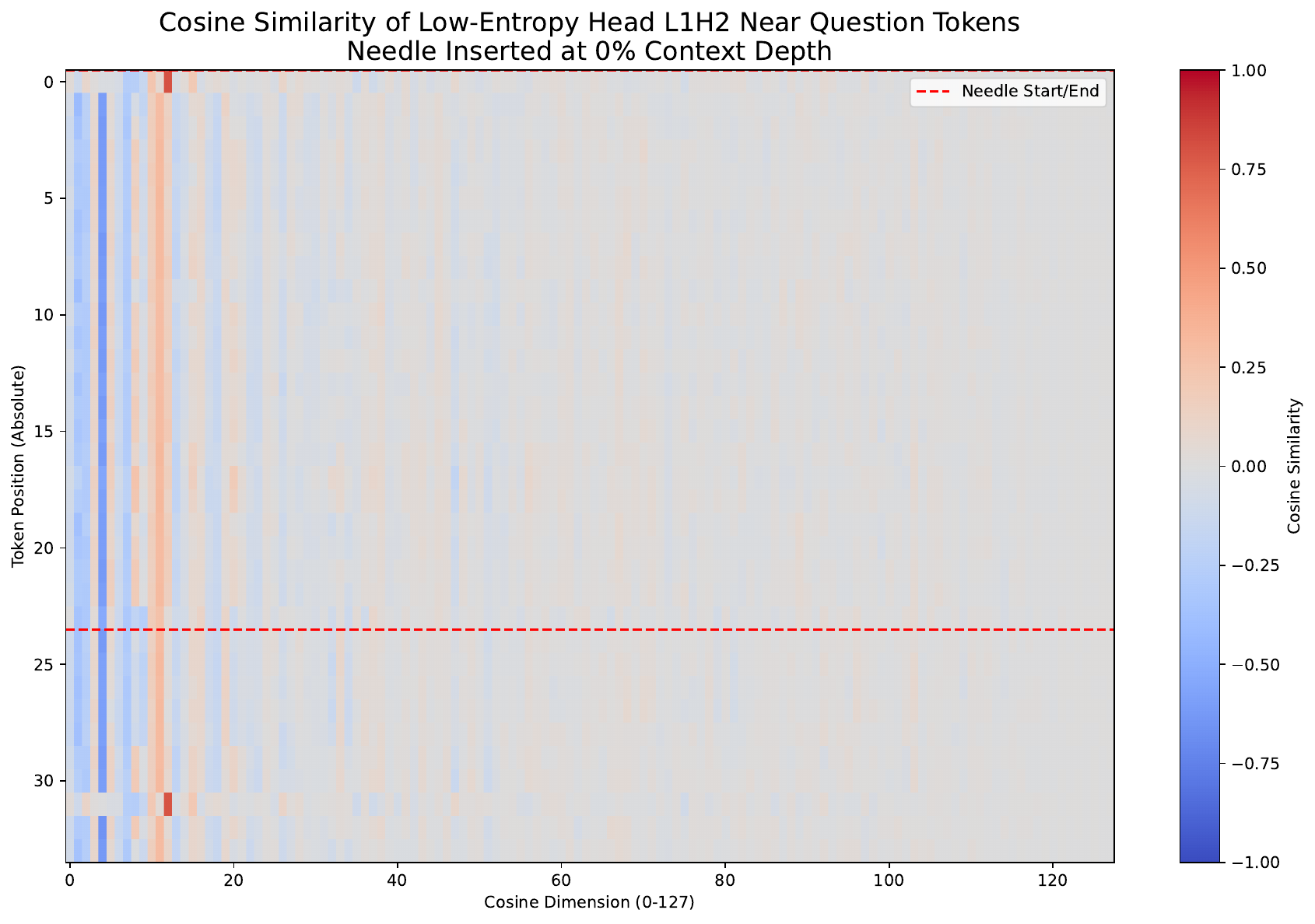}
    \caption{Low matrix entropy head (Layer 1, Head 2)}
    \vspace{-3mm}
    \label{fig:low_entropy_1}
  \end{minipage}
\end{figure*}

\begin{figure*}[!t]
  \centering
  \begin{minipage}{0.48\textwidth}
    \centering
    \includegraphics[width=\linewidth]{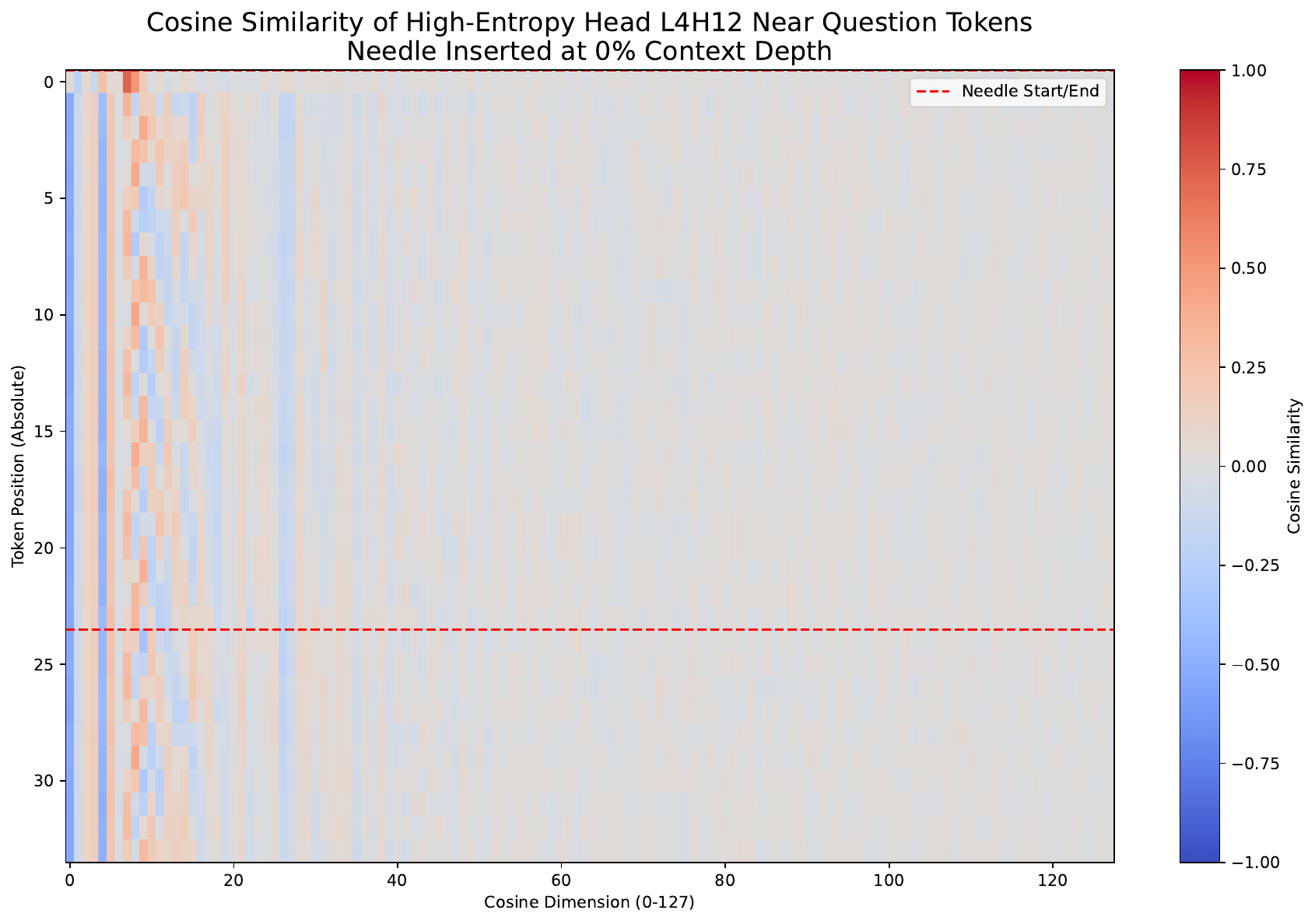}
    \caption{High truncated matrix entropy head (Layer 4, Head 12)}
    \vspace{-4mm}
    \label{fig:high_entropy_2}
  \end{minipage}
  \hfill
  \begin{minipage}{0.48\textwidth}
    \centering
    \includegraphics[width=\linewidth]{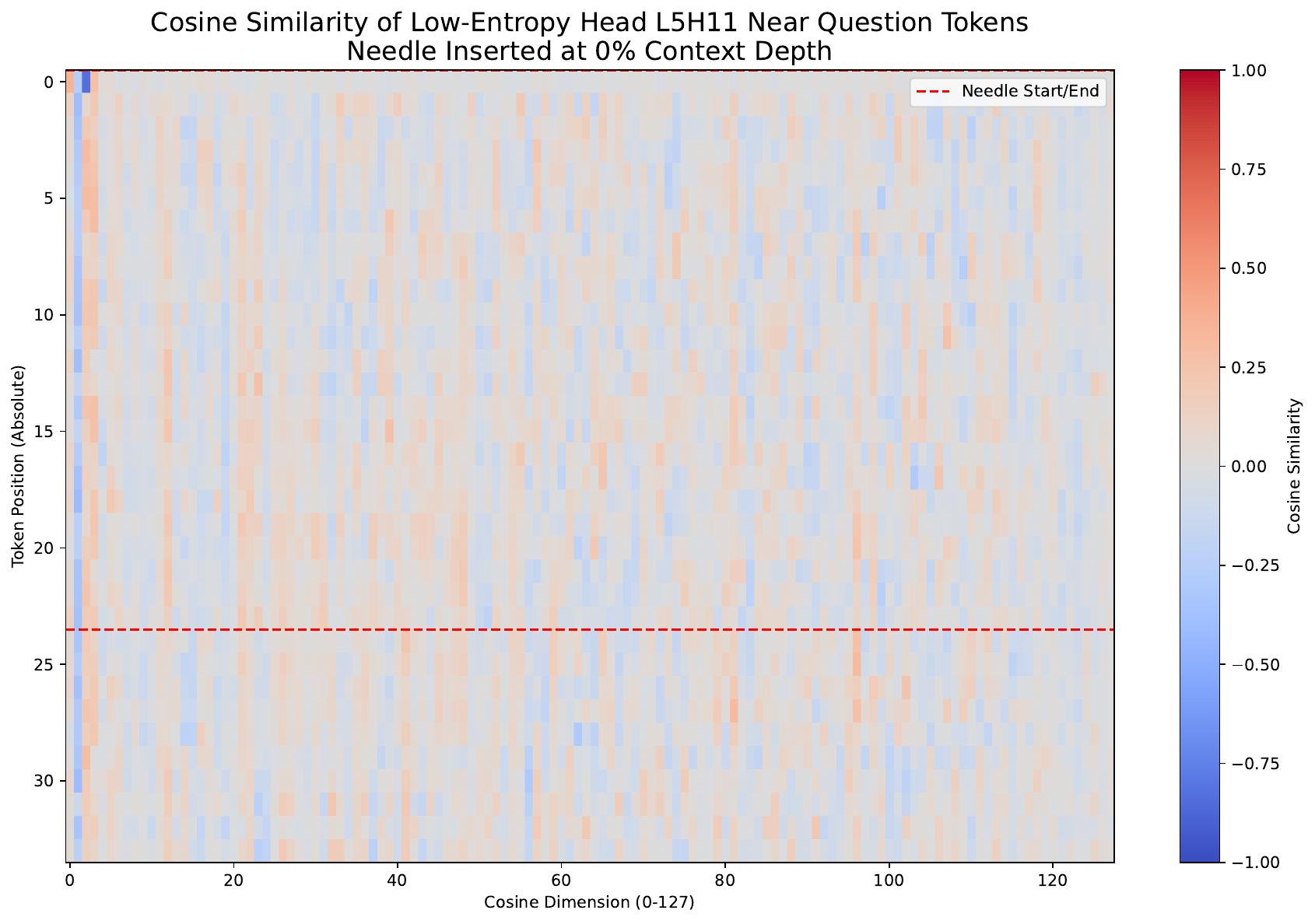}
    \caption{Low truncated matrix entropy head (Layer 5, Head 11)}
    \vspace{-4mm}
    \label{fig:low_entropy_2}
  \end{minipage}
\end{figure*}

\vspace{-3mm}
\subsection{Many-shot In-Context Learning}

We reported the model's performance under many-shot in-context learning (MICL)~\citep{agarwal2024many} in Table~\ref{tab:qwen_math_extrapolation}. We evaluated two settings: inserting the test exemplar into the in-context exemplars (a NIH variant) and omitting the test exemplar (standard in-context learning). Beyond retrieval, this task also probed whether the model extracted reusable reasoning patterns from extended contexts rather than relying on shallow heuristics.

\vspace{-1mm}
\paragraph{The Curse of Length.} We observed that MICL improved reasoning at appropriate lengths, but performance dropped markedly when the context window extended to 16K. Adding more exemplars did not yield further gains, suggesting that learning in ultra-long contexts remained challenging.

\vspace{-1mm}
\paragraph{The Curse of Shortcut.} When we inserted exemplars of the test samples into the in-context examples, we unexpectedly observed a substantial performance drop at 24K and 64K. Instead of copying the correct answers in a ``needle-in-a-haystack'' manner, the model appeared to fall back on shortcuts that hurt overall accuracy.
\vspace{-1mm}
\paragraph{Baseline Failure.}
The two training-free length extrapolation baselines, \emph{Dual Chunk Attention} and \emph{Positional Interpolation}, shown in Table~\ref{tab:qwen_math_extrapolation}, achieve accuracies close to zero, demonstrating astonishingly poor performance.

\vspace{-1mm}
\paragraph{Cross-task Generalization.}
To evaluate cross-task generalization, we compare head selection using entropy computed on either the MATH dataset or the NIH dataset (Table~\ref{tab:dataset_ablation}), and finally test under the MICL task. We find that sparse patterns estimated from synthetic tasks transfer to more complex reasoning tasks, indicating that even without calibration data for the target task, synthetic data still provide useful estimates of the sparse patterns.

\begin{table}[h]
\centering
\caption{Ablation on attention-head identification. We select denoising heads using MATH vs.\ NIH calibration data and evaluate MICL on Qwen2.5-Math-7B at 8K context length. Head selection uses Query representations with the \emph{post-NTK} criterion. Results report MATH accuracy for Needle Insertion and Skip Needle.}
\vspace{-2mm}
\label{tab:dataset_ablation}
\setlength{\tabcolsep}{4pt}%
\renewcommand{\arraystretch}{1.1}%
\begin{adjustbox}{max width=\columnwidth} 
\begin{tabular}{l c c c c c c c}
\toprule
\textbf{Method} & \textbf{Entropy Type} & \textbf{\# Heads} & \textbf{Selection Dataset} & \textbf{Needle Insert (8K)} & \textbf{Skip Needle (8K)}  \\
\midrule
Zero-shot Baseline & -- & -- & -- & 0.430 & 0.430 \\
Many-shot Baseline & -- & -- & -- & 0.240 & 0.230 \\
\midrule
\multicolumn{6}{l}{\textit{Heads selected using MATH dataset}} \\
\midrule
\textsc{DoPE}-by-Gaussian & Trunc-4     & 5 & MATH & 0.375& 0.440 \\
\textsc{DoPE}-by-Gaussian & Trunc-1     & 1 & MATH & 0.393 & 0.410 \\
\textsc{DoPE}-by-parts      & Trunc-4 & 3 & MATH & 0.360 & \textbf{0.420} \\
\textsc{DoPE}-by-parts      & Trunc-1     & 1 & MATH & 0.388 & 0.410 \\
\textsc{DoPE}-by-all     & Trunc-1 & 5 & MATH & 0.338 & \textbf{0.480} \\
\textsc{DoPE}-by-all    & Trunc-1  & 1 & MATH & \textbf{0.395}  & 0.430 \\
\midrule
\multicolumn{6}{l}{\textit{Heads selected using NIH dataset}} \\
\midrule
\textsc{DoPE}-by-Gaussian & Full     & 3 & NIH & 0.365 & 0.390 \\
\textsc{DoPE}-by-Gaussian & Trunc-16     & 1 & NIH & 0.375 & 0.410 \\
\textsc{DoPE}-by-parts     & Trunc-16 & 2 & NIH & 0.372 & 0.330 \\
\textsc{DoPE}-by-parts     & Full     & 2 & NIH & 0.350 & 0.390 \\
\textsc{DoPE}-by-all    & Trunc-16 & 5 & NIH & 0.360 & 0.420 \\
\textsc{DoPE}-by-all    & Trunc-1  & 2 & NIH & \textbf{0.417} & 0.390 \\
\bottomrule
\end{tabular}
\end{adjustbox}
\end{table}
\vspace{-3mm}

\vspace{-3mm}
\subsection{Matrix Entropy Meets Attention Sink}
\vspace{-1mm}
To connect our entropy criterion to attention behavior, we visualized attention distributions for heads identified by high \emph{truncated matrix entropy}. As shown in Fig.~\ref{fig:attn_compare}, \emph{truncated matrix entropy} aligned closely with the \textit{attention sink} phenomenon.

In Fig.~\ref{fig:attn_top16_head}, we observed that when \emph{truncated matrix entropy} identified low-entropy heads, these heads often produced attention sinks (recency bias), while the remaining high-entropy heads allocated attention to the needle. In contrast, Fig.~\ref{fig:attn_all_head} showed that high \emph{vanilla matrix entropy} corresponded to severe attention sinks: although the low-entropy heads displayed more regular attention patterns, they still failed to attend to the needle position.

\vspace{-1mm}
\subsection{RoPE Induces Low-rankness}
\vspace{-2mm}
\label{sec:appendix_visual_analysis}

We then examined how these entropy-based selections related to representation structure by visualizing the token matrix and its corresponding eigenvectors for the heads identified by \textit{Vanilla Matrix Entropy} and \textit{Truncated Matrix Entropy}. Fig.~\ref{fig:high_entropy_1} and \ref{fig:low_entropy_2} showed the cosine similarity between token query vectors (Y-axis, token position) and eigenvectors spanning the \textit{full} 128-dimensional space (X-axis). This projection onto a $k=128$ basis revealed the effective dimensionality used by each head, and highlighted that the two metrics selected different heads with distinct low-rank structure.

\vspace{-2mm}
\paragraph{Low-rankness.}
We observed clear low-rank structure in heads selected by both entropy metrics. Fig.~\ref{fig:low_entropy_1} and Fig.~\ref{fig:high_entropy_2} illustrated heads retained due to \textit{low} Vanilla Matrix Entropy and \textit{high} Truncated Matrix Entropy, respectively. Visually, the similarity mass concentrated in the first few dimensions, indicating that these ``low-rank'' heads relied on only a small subspace to support extrapolation.

\vspace{-2mm}
\paragraph{Periodicity.}
We also observed that the head selected by \emph{truncated matrix entropy} exhibited clear periodicity along the sequence dimension (Fig.~\ref{fig:high_entropy_2}), compared to the head selected by \emph{matrix entropy} in Fig.~\ref{fig:low_entropy_1}. This difference helped explain why \emph{truncated matrix entropy} better identified extrapolative heads: it captured periodic low-rank structure that vanilla matrix entropy did not reliably distinguish.



\vspace{-2mm}
\section{Conclusion}
\vspace{-2mm}
In this paper, we examined how positional encoding shapes long-context behavior in LLMs, with an emphasis on the emergence of \emph{massive activation} and the \emph{attention sink} phenomenon. A key takeaway is that RoPE is not merely a benign carrier of relative position: its low-frequency components can also act as a structured amplifier, concentrating energy and promoting over-aligned, low-rank attention patterns that undermine stability as context grows. \textsc{DoPE} follows directly from this interpretation: rather than modifying RoPE globally or removing positional encoding altogether, it treats instability as a head-specific effect and intervenes only where the attention map becomes noise-dominated. The resulting gains under perturbation suggest that long-context robustness is less about choosing a single positional encoding scheme and more about \emph{controlling} how positional information is injected across heads. 

\section*{Limitations}
\textsc{DoPE} has practical limitations. Head selection adds computation, and the approach assumes that entropy-based criteria reliably separate noise-dominated heads, which may not hold across all models, layers, or domains. Moreover, our evaluation focuses on long-context inference tasks, so generalization to broader settings remains to be validated.

\bibliography{latex/custom}
\clearpage

\appendix

\section{Theoretical Analysis of Spectral Amplification}
\label{sec:appendix}

\subsection{Proofs}
\label{sec:proof}

\begin{lemma}[Entry-level lower bound (rectangular)]
\label{lem:entry-lower-rect}
Let $\mathbf{\Sigma}\in\mathbb{R}^{m\times n}$ with largest singular value $\sigma_1(\mathbf{\Sigma})$. Then
\begin{equation}
\max_{i,j}\big|(\mathbf{\Sigma})_{ij}\big|
\;\ge\;\frac{\sigma_1(\mathbf{\Sigma})}{\sqrt{mn}}.
\end{equation}
\end{lemma}
\noindent
Here $i\!\in\!\{1,\dots,m\}$ and $j\!\in\!\{1,\dots,n\}$ index the rows and columns of $\mathbf{\Sigma}$, respectively.

\begin{proof}
By the Frobenius/spectral norm relation,
{\small
\begin{equation}
\resizebox{\linewidth}{!}{$
\begin{aligned}
\|\mathbf{\Sigma}\|_F^2
&= \sum_{i=1}^{m}\sum_{j=1}^{n} (\mathbf{\Sigma}_{ij})^2
\;\le\; \big(\max_{i,j}|(\mathbf{\Sigma}_{ij})|\big)^2 \, mn,\\
\Rightarrow\qquad
\max_{i,j}|(\mathbf{\Sigma})_{ij}|
&\ge \frac{\|\mathbf{\Sigma}\|_F}{\sqrt{mn}}
\;\ge\; \frac{\sigma_1(\mathbf{\Sigma})}{\sqrt{mn}},
\end{aligned}
$}
\end{equation}
}
since $\|\mathbf{\Sigma}\|_F^2=\sum_r \sigma_r(\mathbf{\Sigma})^2 \ge \sigma_1(\mathbf{\Sigma})^2$.
\end{proof}

\begin{remark}
In the square case $m=n=N$, Lemma~\ref{lem:entry-lower-rect} reduces to
$\max_{i,j}|(\mathbf{\Sigma})_{ij}|\ge \sigma_1(\mathbf{\Sigma})/N$,
where $N$ denotes the sequence length when $\mathbf{\Sigma}$ is formed over $N$ tokens.
\end{remark}

\vspace{1mm}
\subsubsection*{A. Massive Activation in Band-wise Representations}

\begin{lemma}[Cone Condition Implies Coherent Summation]
\label{lem:cone-summation}
Let $\{\widehat{k}_j\}_{j=1}^{N}$ denote the key vectors projected onto a single RoPE frequency band, with
\[
\widehat{k}_j = \beta_j\,\mathbf{R}(\theta_{j,f})\,k,
\qquad
\beta_j \ge \beta_{\min}>0,
\]
where $\mathbf{R}(\theta_{j,f})\!\in\!\mathbb{R}^{2\times2}$ rotates by phase $\theta_{j,f}\!=\!\omega_f j$, and $k$ is a fixed unit vector.  
Assume the \emph{cone condition}: there exists a unit vector $u_f$ and half–angle $\gamma_K<\tfrac{\pi}{2}$ such that
\begin{equation}
\langle u_f,\,\mathbf{R}(\theta_{j,f})k\rangle \;\ge\; \|k\|\cos\gamma_K
\qquad \text{for all } j.
\end{equation}
Then the band-wise sum $S=\sum_{j=1}^{N}\widehat{k}_j$ satisfies
\begin{equation}
\|S\| \;\ge\; N\,\beta_{\min}\,\|k\|\,\cos\gamma_K.
\end{equation}
\end{lemma}
\noindent
Here $N$ is the sequence length (number of token positions in this frequency band).

\begin{proof}
Align $u_f$ with the mean direction of the projected keys.  
Then by linearity and the cone condition,
\begin{equation}
\langle u_f,\,\widehat{k}_j\rangle
= \beta_j\,\langle u_f, \mathbf{R}(\theta_{j,f})k\rangle
\ge \beta_j\,\|k\|\cos\gamma_K.
\end{equation}
Summing over $j$ yields
\begin{equation}
\langle u_f, S\rangle
\ge \|k\|\cos\gamma_K\sum_{j=1}^{N}\beta_j
\ge N\,\beta_{\min}\,\|k\|\cos\gamma_K.
\end{equation}
Since $\|S\|\ge \langle u_f,S\rangle$ for any unit $u_f$, the result follows.
\end{proof}

\begin{theorem}[Spectral Amplification and Massive Activation]
\label{thm:massive-activation}
Under the cone condition of Lemma~\ref{lem:cone-summation}, define the band-wise Gram matrix
\[
\mathbf{\Sigma}_{f}
= \sum_{j=1}^{N}\widehat{k}_j\,\widehat{k}_j^{\!\top}.
\]
Then
\begin{equation}
\lambda_{\max}(\mathbf{\Sigma}_{f})
\;\ge\;
N\,\beta_{\min}^{2}\,\|k\|^{2}\cos^{2}\gamma_K,
\end{equation}
and consequently
\begin{equation}
\sigma_{1}(\mathbf{K}^{\mathrm{R}}_{f})
\;\ge\;
\beta_{\min}\,\|k\|\,\sqrt{N}\,\cos\gamma_K.
\end{equation}
Similarly, for $\mathbf{Q}^{\mathrm{R}}_{f}$ satisfying an analogous cone condition with $(\alpha_{\min},\gamma_Q)$,
\begin{equation}
\sigma_{1}(\mathbf{Q}^{\mathrm{R}}_{f})
\;\ge\;
\alpha_{\min}\,\|q\|\,\sqrt{N}\,\cos\gamma_Q.
\end{equation}
\end{theorem}

\begin{proof}
By the Rayleigh quotient,
\begin{multline}
\lambda_{\max}(\mathbf{\Sigma}_{f})
\ge x^{\!\top}\mathbf{\Sigma}_{f}x
= \sum_{j=1}^{N}(\langle x,\widehat{k}_j\rangle)^2
\\[2pt]
\ge \frac{1}{N}\Big(\sum_{j=1}^{N}\langle x,\widehat{k}_j\rangle\Big)^2
= \frac{\|S\|^2}{N}.
\end{multline}
Applying Lemma~\ref{lem:cone-summation} gives $\lambda_{\max}(\mathbf{\Sigma}_{f})
\ge N\,\beta_{\min}^2\|k\|^2\cos^2\gamma_K$, and taking square roots yields the bound on $\sigma_{1}(\mathbf{K}^{\mathrm{R}}_{f})$.  
Repeating for $\mathbf{Q}^{\mathrm{R}}_{f}$ gives the symmetric result.
\end{proof}

\paragraph{Discussion.}
This theorem shows that within low-frequency RoPE bands, coherent phase rotations accumulate along depth and sequence length, producing $\ell_2$-norm amplification proportional to $\sqrt{N}$—the hallmark of \emph{massive activations} observed in RoPE-based transformers.

\vspace{2mm}
\subsubsection*{B. Attention Sink Amplification in Band-wise Attention Logits}

\begin{theorem}[Spectral Amplification of Attention Scores]
\label{thm:attention-sink}
Given the bounds in Theorem~\ref{thm:massive-activation}, consider the attention submatrix contributed by band $f$,
\begin{equation}
\mathbf{A}_{f}
= \frac{\mathbf{Q}^{\mathrm{R}}_{f}\,\mathbf{K}^{\mathrm{R}}_{f}{}^{\!\top}}{\sqrt{d_h}}.
\end{equation}
Let $\psi$ denote the angle between the dominant singular directions of $\mathbf{Q}^{\mathrm{R}}_{f}$ and $\mathbf{K}^{\mathrm{R}}_{f}$. 
Then its leading singular value satisfies

\begin{equation}
\resizebox{0.87\linewidth}{!}{$
\sigma_{1}(\mathbf{A}_{f})
\;\gtrsim\;
\frac{\alpha_{\min}\beta_{\min}}{\sqrt{d_h}}\,
N\,\|q\|\,\|k\|\,
\cos\gamma_Q\,\cos\gamma_K\,\cos\psi.$}
\end{equation}
Consequently, by Lemma~\ref{lem:entry-lower-rect},
\begin{equation}
\resizebox{0.87\linewidth}{!}{$
\max_{i,j}\big|(\mathbf{A}_f)_{ij}\big|
\;\ge\;
\frac{\alpha_{\min}\beta_{\min}}{\sqrt{d_h}}\,
\|q\|\,\|k\|\,
\cos\gamma_Q\,\cos\gamma_K\,\cos\psi,$}
\end{equation}
which remains $\Omega(1)$ even as sequence length grows.
\end{theorem}

\begin{proof}
Since $\sigma_1(\mathbf{A}_f)\le 
\|\mathbf{Q}^{\mathrm{R}}_{f}\|_2\,\|\mathbf{K}^{\mathrm{R}}_{f}\|_2/\sqrt{d_h}$,
using Theorem~\ref{thm:massive-activation} gives
\begin{equation}
\sigma_{1}(\mathbf{A}_{f})
\gtrsim
\frac{1}{\sqrt{d_h}}\,
\sigma_{1}(\mathbf{Q}^{\mathrm{R}}_{f})\,\sigma_{1}(\mathbf{K}^{\mathrm{R}}_{f})\,\cos\psi,
\end{equation}
yielding the desired bound. 
The entry-level lower bound follows by applying Lemma~\ref{lem:entry-lower-rect}.
\end{proof}

\paragraph{Discussion.}
This second bound extends the massive-activation effect from hidden representations to attention logits.  
When both query and key bands satisfy the cone constraint, their coherent multiplication produces an $\mathcal{O}(N)$ scaled singular value, concentrating attention mass on a few entries, precisely the phenomenon known as \emph{attention sink}.

\vspace{2mm}
\subsubsection*{C. Truncated Matrix Entropy and Spectral Amplification}

\begin{definition}[Truncated Matrix Entropy]
For a head-level Gram matrix $\mathbf{\Sigma}_{h}$ with eigenvalues
$\lambda_{1}\!\ge\!\lambda_{2}\!\ge\!\cdots\!\ge\!\lambda_{r}\!>\!0$,
the \emph{truncated matrix entropy} of order $r$ is
\begin{equation}
\mathcal{H}^{r}_{h}
\;=\;
\frac{1}{r}\sum_{i=1}^{r}\lambda_i\log\lambda_i,
\qquad
r\!\le\!\mathrm{rank}(\mathbf{\Sigma}_{h}).
\end{equation}
It measures the information-weighted energy concentration of the
top-$r$ singular components within an attention head.
\end{definition}

\begin{theorem}[Spectral Amplification Decreases Truncated Entropy]
\label{thm:tme-amplification}
Let $\mathbf{\Sigma}_{f}$ be the band-wise Gram matrix defined in
Theorem~\ref{thm:massive-activation} with eigenvalues
$\lambda_{1}\!\ge\!\lambda_{2}\!\ge\!\cdots$.  
Suppose the cone condition holds and the dominant eigenvalue obeys
\begin{equation}
\begin{aligned}
\lambda_{1}
&\;\ge\;
N\,\beta_{\min}^{2}\,\|k\|^{2}\cos^{2}\gamma_K, \\[3pt]
\sum_{i>1}\lambda_i
&\;\le\;
(1-\delta)\lambda_{1},
\qquad
\delta\!\in\!(0,1).
\end{aligned}
\end{equation}

Then the truncated matrix entropy of order $r\!\ge\!1$ satisfies
\begin{equation}
\resizebox{0.87\linewidth}{!}{$
\mathcal{H}^{r}_{h}
\;\le\;
\lambda_{1}\log\lambda_{1}
+\frac{1-\delta}{r}\lambda_{1}
\log\!\Big(\frac{(1-\delta)\lambda_{1}}{r-1}\Big),$}
\end{equation}
and therefore decreases monotonically with stronger spectral
amplification (i.e., larger $\lambda_{1}$ or smaller $\delta$).
\end{theorem}

\begin{proof}
Let $\lambda_{1}$ be the amplified mode and distribute the remaining
trace mass $(1-\delta)\lambda_{1}$ equally among the next $(r-1)$
eigenvalues, an entropy-maximizing configuration under the given trace
constraint.  Then
\begin{equation}
\resizebox{0.87\linewidth}{!}{$
\mathcal{H}^{r}_{h}
=\frac{1}{r}\Big[
\lambda_{1}\log\lambda_{1}
+(r-1)\tfrac{(1-\delta)\lambda_{1}}{r-1}
\log\!\tfrac{(1-\delta)\lambda_{1}}{r-1}\Big],$}
\end{equation}
which simplifies to the stated bound.  As $\lambda_{1}$ increases or
$\delta$ decreases, the first term dominates and the total entropy
declines, showing that truncated entropy is inversely related to the
degree of spectral concentration.
\end{proof}

\paragraph{Discussion.}
When RoPE’s low-frequency cone constraint amplifies one dominant
spectral direction (large $\lambda_{1}$) while suppressing others
(small $\lambda_{i>1}$), the truncated entropy
$\mathcal{H}^{r}_{h}$ becomes small.  
Hence, heads with low truncated entropy correspond to those exhibiting
\emph{spectral amplification} and potential \emph{attention sinks}.
This justifies using $\mathcal{H}^{r}_{h}$ as a quantitative criterion
to identify “noisy” heads for denoising in DoPE.

\subsection{Experimental Setup}
\paragraph{Models.} Qwen2.5-Math-7B~\citep{yang2024qwen25mathtechnicalreportmathematical} and LLaMA-3-8B-Instruct~\citep{grattafiori2024llama3} are decoder-only transformer models that employ Rotary Positional Embeddings (RoPE) for encoding positional information. Qwen-1.5-7B is trained with a maximum context length of 32K tokens, while LLaMA-3-8B is trained with a 8K-token context window. To support longer contexts beyond their pre-training limits, we apply RoPE-based extrapolation (e.g., Dynamic-NTK), which rescales RoPE frequencies to improve stability and retrieval performance in extended-context settings.

\paragraph{Hyperparameter.}
All experiments use greedy decoding with temperature set to 0.0 and top-$p$ set to 1.0. For the needle-in-a-haystack (NIH) task on LLaMA-3-8B-Instruct, we set \texttt{max\_new\_tokens} to 50 with stop conditions including newline characters (\texttt{<0x0A>}) and stop token ID 144. For the many-shot in-context learning (MICL) task on Qwen2.5-Math-7B, we set \texttt{max\_new\_tokens} to 2048 with stop sequences \texttt{</s>}, \texttt{<|im\_end|>}, \texttt{<|endoftext|>}, and \texttt{Problem:} to prevent generating additional problems. Context buffers of 200 tokens (NIH) and 2,300 tokens (MICL) are reserved for prompt templates and final questions. All experiments are conducted using SGLang~\citep{zheng2023efficiently} (v0.5.3rc0) with the FlashAttention-3 backend~\citep{shah2024flashattention}. Tensor parallelism is enabled for multi-GPU inference when necessary. CUDA graphs are disabled to support dynamic context lengths. All experiments are conducted on five A100 GPUs. Head selection is performed globally across all $(l \times h)$ attention heads, where $l$ is the number of layers and $h$ is the number of heads per layer (32 layers $\times$ 32 heads $=$ 1,024 total heads for LLaMA-3-8B; 28 layers $\times$ 28 heads $=$ 784 total heads for Qwen2.5-Math-7B).

For RoPE extrapolation, we apply Dynamic-NTK scaling~\citep{emo2023dynamicrope} with the scaling factor computed as $\alpha = L_{\text{target}} / L_{\text{original}}$, where $L_{\text{target}} \in \{24\text{K}, 64\text{K}, 128\text{K}\}$ for NIH experiments and $L_{\text{target}} = 16\text{K}$ for MICL experiments, while $L_{\text{original}}$ corresponds to each model's pre-trained maximum position embeddings (32K for Qwen-1.5-7B, 8K for LLaMA-3-8B-Instruct, and 4K for Qwen2.5-Math-7B). For LLaMA-3, we additionally evaluate NTK-by-parts~\citep{peng2023yarn} with \texttt{low\_freq\_factor}$=1.0$ and \texttt{high\_freq\_factor}$=32.0$. The NIH task uses 10 uniformly spaced depth positions (0\%, 10\%, ..., 100\%) for needle insertion at each context length. The MICL task evaluates 100 sampled problems from the MATH dataset~\citep{hendrycksmath2021}, with needle insertion at four fixed depth positions (0\%, 33\%, 67\%, 100\%, corresponding to beginning, 1/3, 2/3, and end) within the in-context examples, yielding 400 total test configurations.

For \textsc{DoPE}, Gaussian noise is sampled from $\mathcal{N}(0, 1)$ with standard deviation $\sigma = 1.0$, using a fixed random seed (42) to ensure reproducibility. The truncated matrix entropy is computed by retaining the top-$k$ singular values where $k \in \{1, 4, 8, 16, 32\}$, with $k=1$ corresponding to using only the spectral norm $\sigma_{\max}(\mathbf{\Sigma})$. We also evaluate the full (untruncated) matrix entropy for comparison.

\paragraph{Baselines.}
Several of the baseline models adopt training-free methods to extend effective context length while preserving short-range quality.

\emph{Dynamic NTK}~\citep{emo2023dynamicrope} adjusts the rotary position embedding (RoPE) base with a \emph{length-dependent} scaling factor at decoding time so that the current effective angular frequency remains closer to the pretraining regime even when the sequence exceeds the original context window. 
Compared with static “NTK-aware” scaling, the dynamic variant reduces frequency drift as length grows and improves long-range stability without additional training; it is also frequently combined with other RoPE extensions in practice.

\emph{Dual Chunk Attention}~\citep{an2024training} is a training-free attention scheme that partitions long sequences into manageable chunks and combines \emph{intra-chunk} attention (local fidelity) with \emph{inter-chunk} routing/aggregation (global recall). 
This design preserves token–token interactions within chunks while enabling information flow across distant chunks, scaling models to $100$k$+$ tokens without continual training, and can be composed with RoPE-based extensions such as PI/NTK-aware/YaRN.

\emph{Positional Interpolation}~\citep{chen2023positionalinterp} linearly down-scales the input position indices before applying RoPE so that positions beyond the original window are \emph{interpolated} back into the training range rather than extrapolated. 
This simple modification allows RoPE-based LLMs to reach 32k-context with minimal fine-tuning while maintaining competitive short-context performance and avoiding unstable attention magnitudes that arise in naive extrapolation.

\end{document}